\documentclass[10pt]{llncs}

\usepackage{graphicx}
\usepackage{amssymb}
\usepackage[latin1]{inputenc}
\usepackage{fuzz}

\begin{document}

\title{A Constrained Object Model for Configuration Based Workflow Composition}

\author{Patrick Albert\inst{1}, Laurent Henocque\inst{2}, Mathias Kleiner\inst{1,2}}

\institute{ILOG, Gentilly, France,\{palbert,mkleiner\}@ilog.fr \and 
LSIS, Marseille, France, laurent.henocque@lsis.org
}

\bibliographystyle{splncs}


\maketitle

\begin{abstract}
Automatic or assisted workflow composition is a field of intense research for applications to the world wide web or to business process modeling. Workflow composition is traditionally addressed in various ways, generally via theorem proving techniques. 
Recent research \cite{ICWS05} observed that building a composite workflow bears strong relationships with finite model search, and that some workflow languages can be defined as constrained object metamodels \cite{UML,YAWL}. This lead to consider the viability of applying configuration techniques to this problem, which was proven feasible. 
Constrained based configuration expects a constrained object model as input. The purpose of this document is to formally specify the constrained object model involved in ongoing experiments and research using the Z specification language, and more precisely using some of the definitions in \cite{racsam}.

\end{abstract}


%

\section{Introduction}\label{sec:intro}

We place ourselves in the scope of automatic or computer aided workflow composition, with immediate applications to Business Process Modeling or the Semantic Web. The basic assumptions for composing workflows is that there exists a form of directory listing of elementary workflows that are potential candidates for composition, as well as a directory listing of transformations that are usable to mediate between workflows having ``bitwise" incompatible message type requirements. How and when a proper list of elementary workflows and transformations can be obtained is beyond the scope of this research, and is treated as if it was available to the program from the start.

We also assume that the composition process is goal oriented: a user may list the message types he can possibly input to the system (e.g. credit card number, expiry date, budget, yes/no answer etc...), and the same user may formulate the precise (set of) message(s) that must be output by the system (e.g. a plane ticket reservation electronic confirmation: the ``goal").

A previous work \cite{ICWS05} proved the feasibility of using a configurator program to solve this problem, and presented a constrained object model adequate for this purpose, using the semi formal language UML/OCL. Although it was shown in \cite{ConfigUsingUML} that such a use of UML/OCL is viable, the language is also known as having limitations, notably concerning relational operators. The original contribution of this work is to propose a formal specification of the same constrained object model using the Z specification language. Z was shown suitable for such a usage in \cite{racsam}, via a framework for the Z specification of constrained object models\footnote{Also called Object Oriented Constraint Programs}. This research heads towards the complete formal specification of a constrained object model for workflow composition.

The plan of the article is as follows. The current section \ref{sec:intro} is introductory, and briefly presents the context retained for dealing with workflow composition configuration in Subsection \ref{sec:context}, a brief introduction to configuration in Subsection \ref{sec:configintro}, how composition can be treated as a configuration task in Section \ref{sec:configtocomposition} and related work in \ref{sec:relatedwork}. Section \ref{sec:introZ} briefly introduces Z and the predefined class construct used in the specification. Section \ref{sec:metamodel} presents the specification of a constrained object model combining a metamodel for activities and data type ontologies. 
Section \ref{sec:conclusion} concludes and presents research perspectives.

\subsection{Context for workflow composition}\label{sec:context}
We consider workflows defined using a variant of extended workflow nets, as are UML2 activity diagrams \cite{UML} or the YAWL language \cite{YAWL}. The underlying model is that of colored Petri nets, where messages (tokens) have types. We do not consider however the underlying semantics of the workflow language, but focus on the properties of the corresponding metamodel. Indeed, we treat workflow composition as the process of connecting input and output message flows to preexisting or added workflow items, like fork, join nodes or auxiliary user input handling actions. Hence the only element retained for composition are the structural properties of argument workflows, messages, and transformations. We do not need to emulate workflows in any case, but can however formulate some constraints that to some extent guarantee the viability of the result.

The same general context is envisioned in several research communications \cite{SOD2004,MultiViewpoint04,PMWSC}. We illustrate our central assumptions by considering the rather complex Producer/Shipper composition problem from \cite{PMWSC}. The problem is to compose a valid workflow from a producer workflow and a shipper workflow. One difficulty is that the execution of both workflows must be interleaved. The producer outputs results that must be fed into the shipper so that both "offers" can be aggregated and presented to the user. This inter-connection remains unknown to the user.  
This example is interesting because:
\begin{itemize}
\item both the shipper and the producer make an offer corresponding to the user request, which are aggregated to make a global offer which can be 
accepted or rejected by the user. Note that the shipper needs input data from the producer to build its offer,
\item both the producer and the shipper are specified using partial workflows, and do not simply amount to simple isolated activities,
\item the two partial workflows cannot be executed one after the other, but they must be interleaved, as each one must wait for the other offer to obtain an OfferAcceptance and therefore complete the transaction,
\item the ShipperWorkflow needs a size as input, which can only be obtained by extraction (i.e transformation) on the ProducerOffer,
\item finally, the goal is decomposed into two sub-goals: the producer and the shipper order confirmations.
\end{itemize}

\begin{figure}[htb]
\begin{center}
   \includegraphics[angle=270,width=8cm]{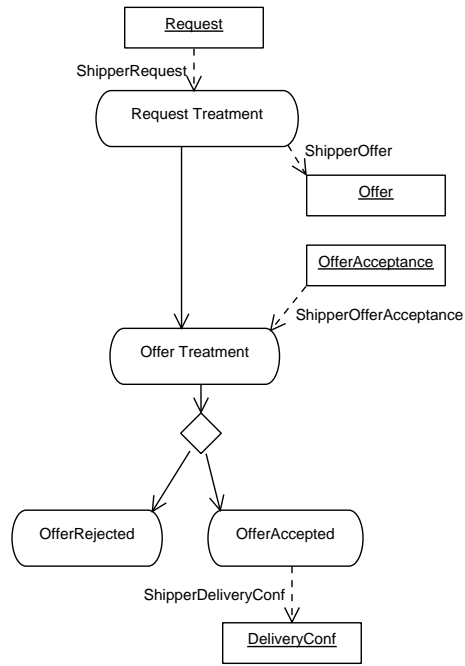}
\vspace{-0,5cm}
\caption{The shipper provided workflow}\label{fig:ShipperWorkflow}
\end{center}
\vspace{-0,5cm}
\end{figure}

Figure \ref{fig:ShipperWorkflow} illustrates the shipper's partial workflow, as defined before search begins. The producer's workflow is similar, modulo the message type ontologies. 

\begin{figure}[htb]
\begin{center}
   \includegraphics[angle=270,width=9cm]{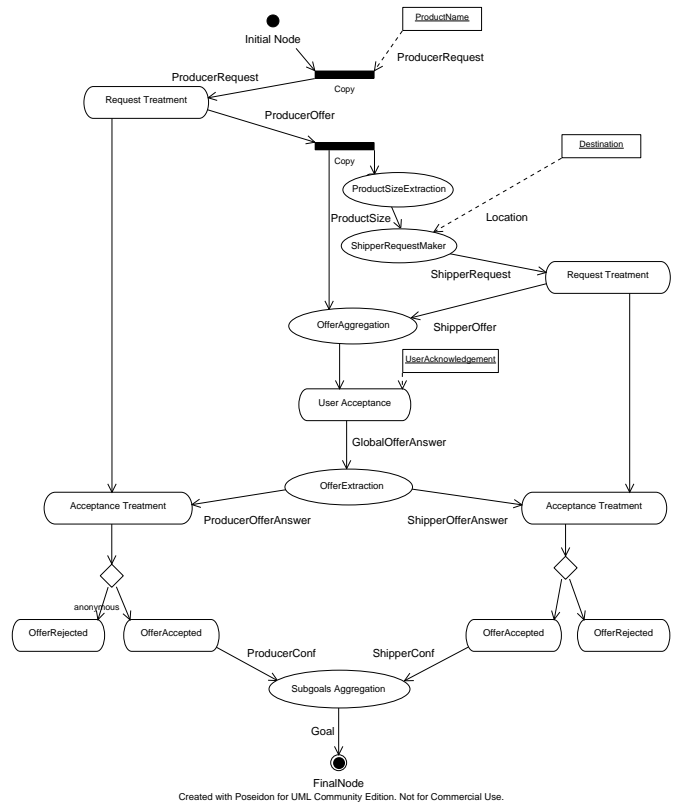}
\caption{The shipper and producer composed workflow}\label{fig:ComposedWorkflow}
\end{center}
\vspace{-0,5cm}
\end{figure}

The composition result is shown in Figure \ref{fig:ComposedWorkflow}. This composed workflow involves synchronization, interleaving, transformations (we used oval boxes to denote transformations) and it should be noted that some execution paths are discarded: indeed, under user rejection, the goal cannot be fulfilled. This illustrates why we later will need a Boolean attribute for active paths in the metamodel. 

Basically, the external user is present via the message it inputs to the global workflow. In Figure \ref{fig:ComposedWorkflow}, all the workflow elements that are not visible in the shipper workflow above or its producer counterpart must be introduced automatically in order to obtain a valid composition. 

\subsection{Brief introduction to configuration}\label{sec:configintro}
A configuration task consists in building (a simulation of) a \emph{complex product} from \emph{components} picked from a catalog of \emph{types}. 
Neither the number nor the actual types of the required components are known beforehand. Components are subject to \emph{relations}, and their types 
are subject to \emph{inheritance}. \emph{Constraints} (also called well-formedness rules) generically define all the valid products. A configurator 
expects as input a fragment of a target object structure, and expands it to a solution of the configuration problem, if any. This problem is semi-decidable in the general case. 

\begin{figure}[htb]
\begin{center}
   \includegraphics[angle=270,width=8cm]{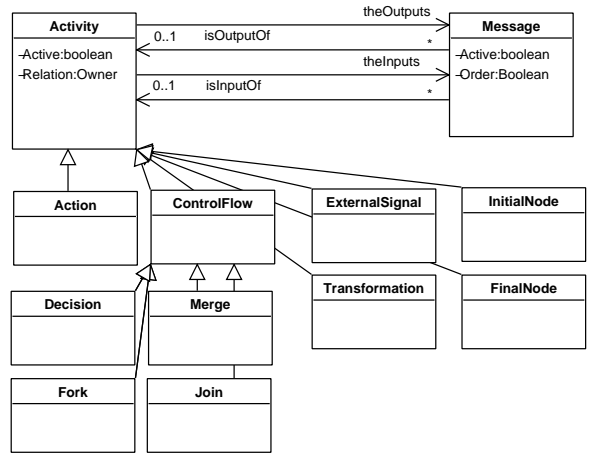}
\caption{Meta-model for workflows activities}\label{fig:modele_activities}
\end{center}
\end{figure}

A configuration program is well described using a \emph{constrained object model} in the form of a standard class diagram (as illustrated by Figures \ref{fig:modele_activities}, \ref{fig:modele_datatypes}), together with well-formedness rules or constraints (also called the semantics in the UML/OCL framework). Technically solving the associated enumeration problem can be made using various formalisms or technical approaches: extensions of the CSP paradigm \cite{Mittal90,Fleischanderl98}, knowledge based approaches \cite{Stumptner97}, terminological logics \cite{Nebel90}, logic programming (using forward or backward chaining, and non standard semantics) \cite{soininen00}, object-oriented approaches \cite{Mailharro98,Stumptner97}. Our experiments were conducted using the object-oriented configurator Ilog JConfigurator \cite{Mailharro98}. 

There currently exists no universally accepted language for specifying constrained object models. The choice of UML/OCL is advocated \cite{ConfigUsingUML}, and is realistic in many situations, but has some drawbacks due to a number of limitations in the OCL language, as for instance the lack of a relational cross product operator. As shown in \cite{racsam} the Z relational language has enough expressive power and extensibility to properly address the task of specifying a constrained object model, without requiring to use an ad hoc object oriented extension of Z.



UML diagrams mention some of the model constraints, most notably relation cardinalities, as e.g. that a message relates to at most one Activity in Figure \ref{fig:modele_activities}, but there is no possibility to graphically cover the whole range of constraints that may occur in an object model.

\subsection{From configuration to workflow composition}\label{sec:configtocomposition}
Configuration emerges as an AI technique with applications in many different areas, where the problem can be formulated as the production of a finite 
instance of an object model subject to constraints. Reasoning about workflows falls into this category, because a workflow description is an instance 
of a given metamodel (as is the UML metamodel for activity diagrams \cite{UML}). Composing workflows is a configuration problem in that in 
so doing, one must introduce an arbitrary number of previously non existent transitions (fork, join, split, merge, transformations, pre-defined 
user-interactions sequences), and interconnect input and output message pins provided they have compatible types. 
The user of a workflow composition system in our sense provides: 
\begin{itemize}
\item a list of potentially usable workflows, implemented in the form of partially defined instances of the workflow metamodel used, (e.g. a producer for some good, a shipper, a payment service, etc.)
\item the ontologies for the data types linked to the input/output messages present in the workflows (e.g. types of journeys: by train, plane, etc. , methods of payment: cheque, credit card, cash etc.)
\item a goal to be satisfied by the result composition (e.g. a train ticket reservation),
\item a list of the data input he can provide (e.g. simple ``yes/no" answers, credit card number, expiry date, selected item etc.) 
\end{itemize} 
In our approach the goal is defined as the type in an appropriate ontology of a message connected to the final node of the composite workflow. 
The inputs provided by the user are modeled as external signals.

The user of a workflow composition system expects in return for his input a complete "\emph{composite}" workflow, that interleaves the execution of several of the elementary argument workflows, while ensuring that all possible integrity constraints remain valid. 
Among such constraints are those that stem from the metamodel itself: for instance some constraints state that two or more workflows should not be inter-blocking, all waiting for some other to send a message. Other constraints are more problem specific, like those stating for instance that an item being shipped is indeed the one that was produced.

\subsection{Related work}\label{sec:relatedwork}
Automated workflow composition is a field of intense activity, with applications to at least two wide areas: Business Process Modeling and the (Semantic) Web Services. Tentative techniques to address this problem are experimented using many formalisms and techniques, among which Situation calculus \cite{adaptinggolog2002}, Logic programming \cite{semiautomatic2003}, Type matching: \cite{largescale2004}, Coloured Petri nets: \cite{SOD2004,MultiViewpoint04}, Linear logic: \cite{linearlogic2004}, Process solving methods \cite{PSMFensel,PSMFramework,ariadne2002}, AI Planning \cite{asplanning2003}, Hierarchical Task Network (HTN) planning \cite{shop2,adaptive2004}, 
Markov decision processes \cite{markov2005}.

\section{Introducing Z}\label{sec:introZ}
\begin{small}
(Note to the reviewer: this section is provided in order to improve the self-containness of the article, and if accepted can be removed from a final version, or moved to an annex, depending upon how tight the page limit is. All relevant references are freely electronically available.)
\end{small}

For space reasons, it is impossible to make this paper self contained, since this would suppose a thorough presentation of both the UML notation \cite{UML}, and the Z specification language \cite{zspivey}. The reader, if novice in these domains, is kindly expected to make his way through the documentation, which is electronically available. 
For clarity however, we provide a brief description of several useful Z constructs. More advanced notations or concepts will be introduced when necessary. 
\subsection{data types as named sets}
Z data types are possibly infinite sets, either uninterpreted ($DATE$), or axiomatically defined as finite sets ($dom$), or declared as explicitly initialized free types ($colors$: :
\begin{zed}
[DATE]
\end{zed}
\vspace{-0,7cm}
\begin{axdef}
dom:\finset \nat
\end{axdef}
\vspace{-0,7cm}
\begin{zed}
	colors ::= red | green | blue
\end{zed}
\noindent From now on, all possible relation types can be built from cross products of other sets. 

\subsection{axiomatic definitions}
Axiomatic definitions allow to define global symbols having plain or relation types. For instance, a finite group is declared as 

\begin{axdef}
zero:dom\\
inverse:dom \fun dom\\
sum:dom \cross dom \fun dom\\
\where
\forall x:dom @ sum(x,inverse(x))=zero\\
\forall x:dom @ sum(x,zero)=x\\
\forall x,y:dom @ sum(x,y)=sum(y,x)\\
\forall x,y,z:dom @ sum(x,sum(y,z))=sum(sum(x,y),z)\\
\end{axdef}

\noindent The previous axiomatic definition illustrates cross products and function definitions as means of typing Z elements. Now axioms or theorems are expressed in classical math style, involving previously defined sets. For instance, we may formulate that the inverse function above is bijective (this is a theorem) in several equivalent ways as e.g.:
\begin{zed}
inverse \in dom \bij dom\\
\end{zed}
where the $\bij$ operator defines a bijection, or explicitly using an appropriate axiom :
\begin{zed}
\forall y:dom@\exists_1 x:dom @ inverse (x)=y
\end{zed}
\subsection{schemas}

\noindent The most important Z construct, \emph{schemas}, occur in the specification in the form of named axiomatic definitions. A schema $[D|P]$ combines one or several variable declarations (in the declaration part $D$) together with a predicate $P$ stating validity conditions (or constraints) that apply to the declared variables.

\begin{schema}{SchemaOne}
a : \nat\\
b : 1 \upto 10
\where
b < a
\end{schema}

\noindent The schema name hides the inner declarations, which are not global. A schema name (as $SchemaOne$ above) is used as a shortcut for its variable and predicate declarations that can be universally or existentially quantified at will. Schemas are $true$ or $false$ under a given \emph{binding}. For instance, $SchemaOne$ is $true$ under the binding $\langle 4 \bind a, 3 \bind b \rangle$ and $false$ under the bindings $\langle 3 \bind a, 234 \bind b \rangle$ or $\langle 3 \bind a, 4 \bind b \rangle$. The latter violates the explicit constraint stated in the predicate part of the schema, while the former also violates the implicit constraint carried by the interval definition $1 \upto 10$ (a subset of $\nat$). In some contexts, a schema name denotes the set of bindings under which it is true.

Z allows Boolean schema composition. Two schemas can be logically combined (e.g. "anded") by merging their declaration parts provided no conflict arises between the types of similarly called variables, and by applying the corresponding logical operator (e.g. the conjunction) to the predicates. For instance, given the schema $SchemaOne$ above, and another schema called $SchemaTwo \defs [b:\nat; c:\nat | b < c]$ \footnote{This illustrates another syntax for simple schema declarations}, 
we may form the schema $SchemaThree$:
\begin{zed}
	SchemaThree \defs SchemaOne \land SchemaTwo
\end{zed}
Incidentally, the variable declarations $b$ in both schemas collide, but not for their types since $b$ is a member of $\nat$ in both cases. The first declaration of $b$ bears a built in constraint, which can be moved to the predicate part. Hence the schema $SchemaThree$ would list as :
\begin{schema}{SchemaThree}
a,b,c : \nat\\
\where
1<b<10\\
b<a\\
b<c
\end{schema}

\subsection{shortcut notation for class specifications}
Z being non object oriented in any way, the specification of an object system is verbose. In \cite{racsam} are proposed the following shortcut definition for classes and types, which makes use of the keywords \emph{class}, \emph{abstract}, \emph{discriminator}, \emph{inherit}. We illustrate here the general framework using a simple three class example. Assume that A,B,C are the sole classes in a constrained object system where B and C inherit A. The Z ``extension" from \cite{racsam} allows for the following simple declarations:  
\begin{schema}{class-A:abstract}
	-discriminators:default\\
	a:\nat
	\where
	a<10;
\end{schema}
\vspace{-0,7cm}
\begin{schema}{class-B:concrete}
	-inherit:A-default\\
	b:\nat_1
	
\end{schema}
\vspace{-0,7cm}
\begin{schema}{class-C:concrete}
	-inherit:A\\
	\where
	a\ge 5;
\end{schema}

\noindent These class declarations are a shortcut for the declaration in the Z specification of diverse sets and axiomatic definitions, of the schemas : $ObjectDef$, $ClassDefA$, $ClassSpecA$, $ClassDefB$, $\dots$, and of the sets $instances(ClassA)$, $A$, $instances(ClassB)$, $\dots$, with:

\begin{zed} [ ObjectReference ] \end{zed}
\vspace{-1cm}

\begin{zed}
    ReferenceSet == \finset ObjectReference
\end{zed}

\noindent Object references are central to the object system, since they allow for specifying object identity. We have three class names:

\begin{zed}
	CLASSNAME ::= ClassA | ClassB | ClassC
\end{zed}

\noindent The function $instances$ maps class names to sets of object references:

\begin{axdef}
instances:CLASSNAME \fun ReferenceSet
\end{axdef}

\noindent The $ObjectDef$ schema introduces a part common to all object representations:

\begin{schema}{ObjectDef}
ref:ObjectReference\\
class:CLASSNAME\\
\end{schema}

\noindent Now, the $ClassDef'X'$ schemas introduce the part specific to each class, plus inheritance using schema inclusion

\begin{schema}{ClassDefA}
a:1 \upto 10
\end{schema}
\vspace{-0,7cm}

\begin{schema}{ClassDefB}
ClassDefA\\
b:\nat_1
\end{schema}
\vspace{-0,7cm}

\begin{schema}{ClassDefC}
ClassDefA \\
\where
a \geq 5\\
\end{schema}

\noindent Class specifications introduce the common $ObjectDef$ part and constrain the $class$ attribute to its proper value:

\begin{zed}
	ClassSpecA \defs ClassDefA \land [ObjectDef | class = ClassA~]\\
	ClassSpecB \defs ClassDefB \land [ObjectDef | class = ClassB~]\\
	ClassSpecC \defs ClassDefC \land [ObjectDef | class = ClassC~]\\
\end{zed}

\noindent Then finally the object system can be modelled, by introducing the sets $A,B,C$ of references that correspond to the usual undestanding of object ``types", again accounting for inheritance: 

\begin{axdef}
A,B,C:ReferenceSet\\
\where
A = instances(ClassA) \cup B \cup C\\
B = instances(ClassB) \\
C = instances(ClassC) \also
instances(ClassA)=\{o : ClassSpecA | o.class=ClassA @ o.i \}\\
instances(ClassB)=\{o : ClassSpecB | o.class=ClassB @ o.i \}\\
instances(ClassC)=\{o : ClassSpecC | o.class=ClassC @ o.i \}\also
\forall i:instances(ClassA) @  (\exists_1 x : ClassSpecA @ x.ref=i)\\
\forall i:instances(ClassB) @  (\exists_1 x : ClassSpecB @ x.ref=i)\\
\forall i:instances(ClassC) @  (\exists_1 x : ClassSpecC @ x.ref=i)\\
\end{axdef}

\noindent The sequel of the presentation makes use of the ``class" shortcuts introduced above, and of some of the role dereferencing operators $\rightarrow$, $\rightharpoonup$, $\cdot$, $\leadsto$ :
\begin{gendef} [X]
\_ \rightarrow \_ : \finset ObjectReference \cross (ObjectReference \fun X) \fun \bag X\\
\_ \rightharpoonup \_ : \finset ObjectReference \cross (ObjectReference \fun X) \fun X\\
\_ \cdot \_ : \finset ObjectReference \cross (\finset ObjectReference \fun \finset X) \fun \finset X\\
\_ \leadsto \_ :  ObjectReference \cross (\finset ObjectReference \fun \finset X) \fun \finset X\\
\where
\forall s:\finset ObjectReference; r: ObjectReference \fun  X @ s \rightarrow r = bagOf(r)(s)\\
\forall s:\finset ObjectReference; r: ObjectReference \fun  X @ \\
\ \ \ \ s \rightharpoonup r = (\mu t: bagOf(r)(s)@first~t)\\
\forall s:\finset ObjectReference; r:\finset ObjectReference \fun \finset X @ s \cdot r = r(s)\\
\forall o:ObjectReference; r:\finset ObjectReference \fun \finset X @ o \leadsto r = r(\{o\})\\
\end{gendef}

\noindent where the function $bagOf$ maps every function from $ObjectReference$ to $X$ to a function from sets of $ObjectReference$ to bags of $X$, assuming the existence of a function $pickFirst$ applied to any set of (totally ordered) object references:

\begin{gendef}[X]
bagOf: (ObjectReference \fun X) \fun (\finset ObjectReference \fun \bag X )
\where
\forall f:ObjectReference \fun X @ bagOf(f)(\emptyset)=\lbag \rbag\\
\forall f:ObjectReference \fun X @ \\
\t1 \forall d:\finset_1 (\dom f)  @ (\LET x == pickFirst(d) @ \\
\t1 bagOf(f)(d) =  (bagOf(f)(d\setminus \{x\}) \uplus (\{f(x) \mapsto 1\})))
\end{gendef}

\section{A metamodel for workflow composition}\label{sec:metamodel}
Workflow reasoning requires a workflow language with enough generality to be practically viable. Furthermore in our case, since we expect to treat 
workflow composition as a configuration task, it is of particular importance that the language is modular wrt. most if not all the workflow patterns 
referenced in \cite{WFPATTERNS}. The simplest such language is the extended workflow net YAWL language \cite{YAWL}, now a subset of UML2 
\cite{UML} activity diagrams. 

We present our constrained object model according to the standard model driven architecture recommendations, except for the use of the Z language as a formal specification language. The next subsection introduces the classes, and their 
associations and attributes using class diagrams. This presentation is then followed by a detailed presentation of the relevant constraints, in the form of Z axiomatic definitions. 

\subsection{Metamodel specification}

\subsubsection{Actions}
Actions are the core constituents of a workflow. As defined in UML2, actions may have a certain number of messages as their inputs and outputs. Those 
inputs/outputs have defined types, taken from existing ontologies of data types. All actions have an "owner" (the original workflow they belong to: for instance, an action may belong to the Producer workflow). In UML2 the term workflow is synonym to that of Activity.
All workflow parts that are dynamically added by the configurator in order to create the composed workflow belong to a newly introduced owner called the \emph{Composition Workflow}. There are different types of actions, illustrated in the metamodel in Figure \ref{fig:modele_activities}:
\begin{itemize}
\item Initial nodes: the starting point of the workflow. Initial nodes don't take any inputs. They are graphically represented using a black circle.
\item Final nodes: a possible end of the workflow. Final nodes don't produce any outputs and are represented using a white circle and a black dot in the center. There may be several final nodes in a workflow.
\item Control nodes: joins, forks, merges, decisions. A fork initiates concurrency by duplicating its input token to all outputs. Join is the corresponding synchronization construct. Decision (also known as ``split") and merge are the standard if/else conditional branching constructs.
\item Actions: activities which include a local action executed by the workflow owner.
\item Transformations: activities for transforming message data types with no further side effect. Transformations are called data mediators in the context of web service composition. Available transformations can be chosen by the composition designer, or they can be discovered (as e.g. in the context of Semantical Web Services). The distinction between actions in general and transformations in particular is widely acknowledged in the workflow/process/WS communities. Called ``data mediators" in the context of WS, UML2 ``transformations" have the sole effect of reformatting data, and do not enter in further interactions with other parts of the workflow.
\item External Signals: An activity that outputs external messages, typically user provided messages.

\end{itemize}
\subsubsection{Class specifications}
We now introduce the Z specification of the constrained object model used for the configuration of workflow compositions. The class specifications listed below straightforwardly follow from  Figure \ref{fig:modele_activities}.
We use the notational shortcuts introduced in \cite{racsam} for class declarations. These shortcuts allow for straightforward class definitions, and introduce several (hidden) auxiliary data types and sets. We also take the freedom of introducing the type "Boolean" in the language, for the sake of simplicity.
\begin{zed}
Boolean::=true|false
\end{zed}
\vspace{-0,7cm}
\begin{schema}{class-Activity:abstract}
	active:Boolean\\
\end{schema}
\vspace{-0,7cm}
\begin{schema}{class-Action:concrete}
	-inherit:Activity\\
\end{schema}
\vspace{-0,7cm}
\begin{schema}{class-ControlFlow:abstract}
	-inherit:Activity\\
\end{schema}
\vspace{-0,7cm}
\begin{schema}{class-ExternalSignal:concrete}
-inherit:Activity
\end{schema}
\noindent and similarly for $InitialNode$, $FinalNode$, $Transformation$.
\begin{schema}{class-Decision:concrete}
	-inherit:ControlFlow\\
\end{schema}
\noindent and similarly for $Merge$, $Split$, $Join$.
\begin{schema}{class-Message:concrete}
	active:Boolean\\
	order:\nat
\where
order \ge 0	
\end{schema}
\vspace{-0,7cm}

\subsubsection{Relation specifications}
According to Figure \ref{fig:modele_activities} there exists a relation between workflows and abstract actions: each action has a single owner workflow. This can be modeled using an injection
\begin{axdef}
owner:  Activity \inj Workflow\\
\end{axdef}
The relations listed in Figure \ref{fig:modele_activities} between the Activity and Message classes can be specified as:
\begin{axdef}
outputs: Activity \rel  Message\\
inputs: Activity \rel  Message\\
isOutputOf: Message \pinj Activity\\
isInputOf: Message \pinj Activity
\where
isOutputOf=outputs \inv\\
isInputOf=inputs \inv
\end{axdef}
where $inputs \inv$ denotes the relational inverse of the relation $inputs$, and $\pinj$ denotes a partial injection. Partial injections are useful to specify situations modeled using $0,1$ cardinalities as in Figure \ref{fig:modele_activities}.

\subsection{Ontology of message types}
\begin{figure}[htb]
\begin{center}
   \includegraphics[angle=270,width=10cm]{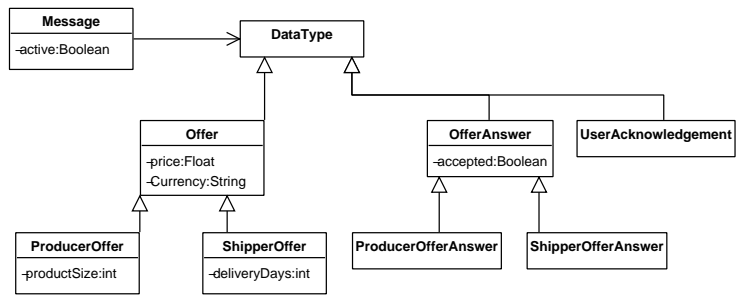}
\caption{Abstract model for workflow data types ontologies}\label{fig:modele_datatypes}
\end{center}
\vspace{-0,5cm}
\end{figure}

Each message has a related data type, from a workflow specific ontology. We use predefined ontologies for user interaction schemes, 
and import the ones required by the selected web services. User interaction schemes, as implemented by the composition activity \emph{OfferAcceptance}, constrain the types of their I/O messages. From an abstract standpoint, they output an \emph{OfferAnswer} if both an \emph{Offer} and an \emph{UserAcknowledgement} are provided. However the precise \emph{Offer}/\emph{OfferAnswer} type match is constrained: they must share the same owner workflow.
For example, a \emph{ShipperOfferAnswer} can be output only if a \emph{ShipperOffer} is input to the user interaction. Figure \ref{fig:modele_datatypes} illustrates the fact that such answers belong to both the hierarchy of standard datatypes and of imported service ontologies.

\subsubsection{Class specifications}
The class specifications straightforwardly follow from Figure \ref{fig:modele_datatypes}:
\begin{zed}
Currency::=Euro | Dollar | Yen \ldots
\end{zed}
\vspace{-0,7cm}
\begin{schema}{class-DataType:abstract}
\end{schema}
\vspace{-0,7cm}
\begin{schema}{class-Offer:abstract}
-inherit:DataType\\
price:\nat\\
currency:Currency
\end{schema}
\vspace{-0,7cm}
\begin{schema}{class-OfferAnswer:abstract}
-inherit:DataType\\
accepted:Boolean
\end{schema}
\vspace{-0,7cm}
\begin{schema}{class-UserAcknowledgement:concrete}
-inherit:DataType
\end{schema}
\vspace{-0,7cm}
\begin{schema}{class-ProducerOffer:concrete}
-inherit:Offer\\
size:\nat
\end{schema}
\vspace{-0,7cm}
\begin{schema}{class-ProducerOfferAnswer:concrete}
-inherit:OfferAnswer
\end{schema}
\vspace{-0,7cm}
\begin{schema}{class-ShipperOffer:concrete}
-inherit:Offer\\
deliveryDays:\nat
\end{schema}
\vspace{-0,7cm}
\begin{schema}{class-ShipperOfferAnswer:concrete}
-inherit:OfferAnswer
\end{schema}
\vspace{-0,7cm}

\subsubsection{Relations}
There is a relation between the Message and DataType class, whereby each Message binds to at most one DataType object. This is specified using a partial function. The same DataType instance might be shared across several Messages, hence the function $dataType$ is not injective

\begin{axdef}
dataType:  Message \pfun DataType\\
\end{axdef}

\subsection{Semantics}
The previous object models are not enough to describe valid compositions, and require a number of constraints governing the possible combinations of partial workflows and additional elements that form acceptable compositions. We specify here a limited number of these constraints that are representative enough to grasp the general idea.

\subsubsection{Composition specific constraints}

From the simplified and slightly adapted subset of the UML2 activity diagram metamodel in Figure \ref{fig:modele_activities}, we observe that both the \emph{Activity} and \emph{Message} classes implement a Boolean attribute called ``active". This Boolean helps ensuring that a workflow can be composed from sub-workflows if and only if at least one valid path yields the expected goal. 
This allows our tool to produce composite workflows under the additional constraint that control flow constructs must match the following constraints applying to activities and messages:
\begin{itemize}
\item if an action is active, then all its input messages are active, 
\item if an active message is output of a join, then all inputs of the join must be active,
\item if an active message is output of a decision or a fork, then the input of this activity must be active,
\item if an active message is output of a merge, at least one of this merge's inputs must be active.
\end{itemize}
According with these, the program builds solutions such that at least one path leads from the initial node to a final node reached via a message having the correct goal type. In the case a valid path traverses a fork or a join, all other incoming/outgoing paths must be valid too. If a user wants a robust solution (meaning that all branches are valid), this can be obtained by forcing all parts of the workflow to be active.
\subsubsection{Activation related constraints}
These constraints are not problem-specific and therefore apply to any composition. The Boolean "active" denotes which part of a workflow indeed participate in the solution. The rationale for this is as follows: a workflow argument to a composition may involve decision/merge paths that are ignored because either the conditions for their activation are known as impossible (e.g. because from the connected message, we know that a test will always fail) or because an exterior message required for their successful execution is known as missing (e.g. a user message giving a credit card number in case the user has none).

\noindent ``If an action is active, then all of its inputs must be active messages":
 
\begin{zed}
\forall a:Action @ a.active \implies \forall m \in inputs(a) @ m.active\\
\end{zed}

\noindent ``If an active message is output of a join, then all inputs of the join must be active":
 
\begin{zed}
\forall m:Message | m.active \wedge m \leadsto isOutputOf \in Join @ \\
\ \ \ \ \forall m':Message | m' \in m \leadsto isOutputOf \leadsto inputs @ m'.active\\
\end{zed}

\noindent ``If an active message is output of a decision or a fork, then all inputs of this activity must be active":
 
\begin{zed}
\forall m:Message | m.active \wedge m \leadsto isOutputOf \in Decision \cup Fork @ \\
\ \ \ \ \forall m':Message | m' \in m \leadsto isOutputOf \leadsto inputs @ m'.active\\
\end{zed}

\noindent ``If an active message is output of a merge, at least one of this merge's inputs must be active":
 
\begin{zed}
\forall m:Message | m.active \wedge m \leadsto isOutputOf \in Merge @ \\
\ \ \ \ \exists m':Message | m' \in m \leadsto isOutputOf \leadsto inputs @ m'.active\\
\end{zed}

\noindent ``If a merge is active, then at least one of its inputs must be an active message":
 
\begin{zed}
\forall a:Merge @ a.active \implies \exists m \in inputs(a) @ m.active\\
\end{zed}

\subsubsection{Composition related constraints}
We assume the existence of a specific workflow instance called ``Composition":
\begin{axdef}
Composition:Workflow
\end{axdef}

\noindent ``All messages input of an external workflow are output of the composition workflow":
 
\begin{zed}
\forall m:Message @ m\leadsto isInputOf \leadsto owner \neq Composition \implies\\
\ \ \ \   m\leadsto isOutputOf \leadsto owner=Composition\\
\end{zed}

\noindent and conversely ``All messages output of an external workflow are input of the composition workflow":

\begin{zed}
\forall m:Message @ m\leadsto isOutputOf \leadsto owner \neq Composition \implies\\
\ \ \ \   m\leadsto isInputOf \leadsto owner=Composition\\
\end{zed}

\subsubsection{Message ordering related constraints }

Our model implements an integer ``order" parameter in the Message class that is used to prevent building interlocking or looping constructs.

\noindent ``the order of an action's input message is lower than the action's output messages orders (this ``ordering" constraint allows to prevent looping situations in the composite workflow):

\begin{zed}
\forall m:Message @ \forall m':Message | \\
\ \ \ \  m' \in m \leadsto isInputOf \leadsto outputs @ m.order < m'.order
\end{zed}
 
\subsubsection{Pre-defined composition constraints}
An OfferAcceptance action expects as input an Offer plus a UserAcknowledgement, and produces an OfferAnswer as a result. Both the Offer and the OfferAnswer point to actions having the same workflow owner, which is not the Composition workflow.

\begin{zed}
\forall o:Offer,a:OfferAnswer | o\leadsto isInputOf=a\leadsto outputOf @\\
\ \ \ \  o\leadsto isOutputOf \leadsto owner = a \leadsto isInputOf \leadsto owner\\
\end{zed}

\noindent Also, Fork nodes have the same type for their inputs and outputs. Formulating such a constraint is possible, under the assumptions in \cite{racsam}
\begin{zed}
\forall f:Fork, i,o:Message|i\leadsto isInputOf=o\leadsto isOutputOf=f@\\
\ \ \ \ i\leadsto getClass=o\leadsto getClass
\end{zed}

\subsubsection{Problem specific constraints}

User provided message types fall into a few categories, as e.g. credit card information, age, or budget...
Assuming the user input message type classes UserInput1, UserInput2, ...
\begin{zed}
\forall m:Message|m\leadsto isOutputOf \in ExternalSignal @\\
\ \ \ \  m \leadsto getClass \in {UserInput1,UserInput2, \ldots} 
\end{zed}

\noindent Also, a concrete composition instance must list the available transformation types. In the context of (semantic) web service discovery, such transformations may be the result of a query to a repository of ontology mediators.
 
\noindent Finally, a precise workflow composition problem instance may involve policy related constraints: constraints that are required to filter out valid yet unwanted compositions, for instance in a way such that offer's prices fall below a given maximum value.


\section{Conclusion}\label{sec:conclusion}
This work proposes a formal specification using the Z language of a constrained object model involved in automatic workflow composition. Constrained object models can be exploited straightforwardly by configurators to achieve automatic or assisted workflow composition. Our formalization abstracts from the technology used to assess the validity of such an approach, so that different configuration techniques can be tested or compared on the same problem.

Configuration expects a constrained object model to operate, hence puts the application design in a field familiar to many engineers. An essential part of the object model, the metamodel for activity diagrams, already exists as a (subset of) part of the UML2 specification relative to activity diagrams. Our Z specification is compatible with all UML2 class diagram features (including multiple inheritance and inheritance discriminators), thus allowing for the straightforward translation of class diagrams. The advantages of Z wrt. UML/OCL are in the statement of constraints. For instance UML dramatically lacks relational constructs and a cross product operator. Since in configuration problems, relations abound with extra semantics (injectivity etc.), object model constraints can be freely stated in Z, also taking advantage of using the richness of Z relational operators, and the ability to define additional operators.


\section*{Acknowledgments}
The authors would like to thank the European DIP integrated project and the ILOG company for their financial support in carrying out this research. This reasearch was partly carried on while the second author was having a research residence at GRIMAAG, University of Antilles Guyane.



%

\bibliography{composition_bpm05}




\end{document}